\documentclass[12pt]{article}
 \usepackage{amssymb}
 \usepackage{amsmath}
\usepackage{tikz}
\usepackage[square,numbers]{natbib}
\usepackage{float}
\usetikzlibrary{matrix}
\usepackage{pgfplots}
\pgfplotsset{
	compat=1.11,
}

\title{%\vspace{-1.5cm}          
	A Gentle Lecture Note on Filtrations\\ in Reinforcement Learning}
\author{W.J.A. van Heeswijk}

\begin{document}
	\maketitle
	
	\begin{abstract}
	This note aims to provide a basic intuition on the concept of filtrations as used in the context of reinforcement learning (RL). Filtrations are often used to formally define  RL problems, yet their implications might not be eminent for those without a background in measure theory. Essentially, a filtration is a construct that captures partial knowledge up to time $t$, without revealing any future information that has already been simulated, yet not revealed to the decision-maker. We illustrate this with simple examples from the finance domain on both discrete and continuous outcome spaces. Furthermore, we show that the notion of filtration is not needed, as basing decisions solely on the current problem state (which is possible due to the Markovian property) suffices to eliminate future knowledge from the decision-making process.
	\end{abstract}
	
When reinforcement learning (RL) problems are introduced, papers typically start with some generic Markov Decision Process (MDP) model that looks something like $(\mathcal{S},\mathcal{X}(S_{t+1}),\mathbb{P}^\Omega(S_{t+1} \mid S_t,x_t),R(S_t,x_t),\rho)$ \cite{vanheeswijk2020}. In this tuple, $\mathcal{S}$ defines the set of all problem states, $\mathcal{X}(S_t)$ describes the set of feasible decisions (given some state $S_t \in \mathcal{S}$), $\mathbb{P}^\Omega(S_{t+1} \mid S_t,x_t)$ is the probability measure on outcome space (also known as sample space) $\Omega$ that describes state transitions (a probability mass function for discrete outcome spaces and a  probability density function for continuous outcome spaces), $R(S_t,x_t)$ is the reward function that computes rewards for a given state-action pair, and $\rho \in (0,1)$ is the discount factor for future rewards. The outcome space $\Omega$ includes all possible events that may occur, with $\omega \in \Omega$ representing a particular realization of an event path (or sample path) in the outcome space. This paper assumes a finite time horizon $\mathcal{T}=\{0,1,\ldots,T\}$, in this case we may define $\omega=\{\omega_1,\omega_2,\ldots,\omega_T\}$ as the ordered set of events, which combined with the initial state $S_0$ and the sequence of decisions enables to compute all states that are visited. Finally, the Markovian property -- also known as memoryless property -- by definition holds for any MDP, meaning that the probability measure $\mathbb{P}^\Omega$ is conditional only on the present state, not on states and events in the past \cite{powell2020}.

Reinforcement learning aims to approximately solve MDP models and find a decision-making policy $\pi: S_t \mapsto x_t$. Whatever flavor of RL is used, at the framework's core Monte Carlo simulation is performed to repeatedly sample paths in the outcome space and learn good decisions based on these observed paths. In line with $\mathbb{P}^\Omega$ we sample random variables $W_t$ with realizations $W_t=\omega_t$.

In addition to the aforementioned model conventions, it is often mentioned that the decision-making policy is $\mathcal{F}_t$-measurable, that we deal with a filtered probability space, or that the expected value is conditional on a filtration; this notion of `filtration' originates from the field of measure theory. Particularly for RL researchers from more applied backgrounds, the implications of a filtered probability space might not be eminent. When looking up the corresponding textbook definition of filtrations (see, e.g., \cite{shiryaev1996,shreve2004}), you will probably find something like this:

\begin{quote} 	
\textit{Let $(W_1,W_2,\ldots,W_T)$ be the sequence of information variables defined over $\mathcal{T}$, containing an ordered set of exogenous information $W_t$. Let $\omega \in \Omega$ be a sample sequence of an event realization  $W_1=\omega_1,W_2=\omega_2,\ldots,W_T=\omega_T$. Furthermore, let $\mathcal{F}$ be the $\sigma$-algebra on $\Omega$, capturing all possible events included in $\Omega$. The set $\mathcal{F}$ is composed of all countable unions and complements of the elements defined in $\Omega$. Let $\mathbb{P}^\Omega$ be a probability measure on $(\Omega,\mathcal{F})$. Let $\mathcal{F}_t = \sigma(W_1,\ldots,W_t)$ be the $\sigma$-algebra generated by the process $(W_1,\ldots,W_t)$, containing all subsets of $\Omega$ conditional on the information sequence that has been revealed up to time $t$. The sequence $\mathcal{F}_0,\mathcal{F}_1,\ldots,\mathcal{F}_t$ is a filtration that is subject to $\mathcal{F}_t \subseteq \mathcal{F}_{t+1}, \forall t \in \mathcal{T}$.}
\end{quote}

Although such introductions are needed to rigorously define the concept, they do not necessarily offer an intuitive understanding. We therefore provide a (hopefully) more intuitive background, followed by a toy-sized example. From a mathematical perspective, the outcome space $\Omega$ is simply a set containing elements $\omega$. In an RL context the term `sample space' is often more appropriate, as we randomly sample outcomes while simulating time transitions. Examples of the outcome space might be: the attainable outcomes of the cast of a die, the possible movements of a stock price, potential arrivals of new jobs, etc. At this point, it is appropriate to define the event $A \in \Omega$, which for convenience we may think of as a set of outcomes with a corresponding `yes' answer or some common property. The complementary set $A^C$ is the set where the answer is `no' or the property is absent. Each event has a positive probability that we can measure, e.g., the probability that a random number falls within a certain interval.

A filtration essentially is a mathematical model that represents partial knowledge about the outcome. Intuitively, if $\mathcal{F}_t$ is the filtration and $A \in \Omega$, then if $A \in \mathcal{F}_t$ we know whether $\omega \in A$ or not. In plain words: the filtration tells us whether an event happened or not. Furthermore, the filtration expands with the passing of time, as indicated by the property $\mathcal{F}_t \subseteq \mathcal{F}_{t+1}$. One may envision the `filtration process' as a sequence of filters, each filter providing us a more detailed view of the events in $\Omega$. In the context of MDPs and RL, a filtration $\mathcal{F}_t$ provides us with the necessary information to compute the current state $S_t$. At the same time, the information embedded in the filtration cannot give any indication of future changes in the process \cite{shreve2004}. We know the event path up to $t$, but not the events that will occur after that. Observe that this observation coincides with the Markovian property.

As a filtration is a $\sigma$-algebra, a basic understanding of $\sigma$-algebras is essential, although we need not to discuss them in great detail. Loosely defined, a $\sigma$-algebra is a collection of subsets of the outcome space, containing a countable number of events as well as all their complements and unions. Essentially, the $\sigma$-algebra allows to define certain measures (e.g., length, volume), which would not be possible for every subset of the outcome space. 
 
  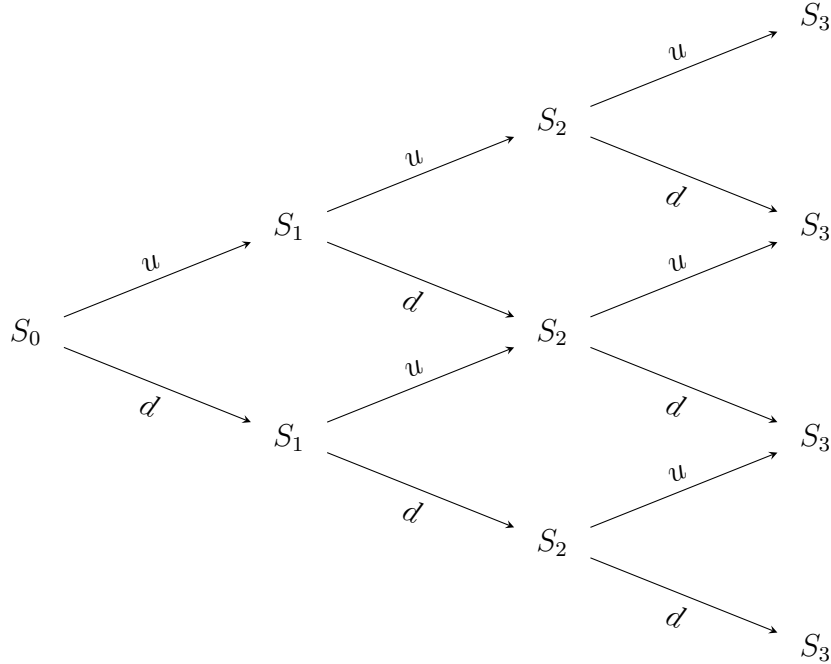
\begin{figure}[H]
  	\centering
  	\begin{tikzpicture}[>=stealth,sloped]
  	\matrix (tree) [%
  	matrix of nodes,
  	minimum size=1cm,
  	column sep=2.5cm,
  	row sep=0.4cm,
  	]
   	{  
   		&         &         &  $S_3$             \\
   		&         &  $S_2$  &                  \\
   		& $S_1$   &         &  $S_3$             \\
   		$ S_0$ &         &  $S_2$  &                  \\
   		& $S_1$   &         &  $S_3$         \\
   		&         &  $S_2$  &                \\
   		&         &         &  $S_3$    \\
   	};
  	\draw[->] (tree-4-1) -- (tree-3-2) node [midway,above] {$u$};
  	\draw[->] (tree-4-1) -- (tree-5-2) node [midway,below] {$d$};
  	\draw[->] (tree-3-2) -- (tree-2-3) node [midway,above] {$u$};
  	\draw[->] (tree-3-2) -- (tree-4-3) node [midway,below] {$d$};
  	\draw[->] (tree-5-2) -- (tree-4-3) node [midway,above] {$u$};
  	\draw[->] (tree-5-2) -- (tree-6-3) node [midway,below] {$d$};
  	
  	 	\draw[->] (tree-2-3) -- (tree-1-4) node [midway,above] {$u$};
  		\draw[->] (tree-2-3) -- (tree-3-4) node [midway,below] {$d$};
  	 	\draw[->] (tree-4-3) -- (tree-3-4) node [midway,above] {$u$};
  	 	\draw[->] (tree-4-3) -- (tree-5-4) node [midway,below] {$d$};
  	 	\draw[->] (tree-6-3) -- (tree-5-4) node [midway,above] {$u$};
  	 	\draw[->] (tree-6-3) -- (tree-7-4) node [midway,below] {$d$};
  	\end{tikzpicture}
  	\caption{Binomial lattice model with $T=3$. With each time step, the stock price $S_t$ goes either up ($u$) or down ($d$).}
  	\label{fig:binomiallattice}
  \end{figure}

 %Filtration: what about complements???
 %Typically a filtration at 0 is defined by $\mathcal{F}_0=\emptyset,\Omega$. We use a slightly modified notation that includes the initial state $S_0$; this is often helpful in the context of RL.
 We proceed to introduce an example.  To illustrate the concept of filtration as simply as possible, we introduce a problem setting in which the state $S_t \in \mathbb{R}^+$ represents the price of a given financial stock at time $t$. No other model information is needed for this exercise, although for practical purposes you might imagine that you aim to buy at a low price and sell at a high price to lock in profits. Suppose the initial stock price is defined by $S_0$ and we have a time horizon composed of three discrete time steps ($T=3$). We define a simplified binomial lattice model to reflect price movements: at each time step the stock price can go either up ($u$) or down ($d$). The realizations $u$ and $d$ are added to (subtracted from) the preceding price ($S_1=S_0 + u$ or $S_1=S_0 -d$), for details on binomial lattices we refer the interested reader to \cite{luenberger1997}. In terms of samples, we have $\omega_t \in \{u,d\}, \forall t \in \mathcal{T}$ and $\omega=\{\omega_1,\omega_2,\omega_3\}$ describes a realization of a price path, e.g., $\omega=\{u,d,u\}$. The binomial lattice corresponding to the example is depicted in Figure~\ref{fig:binomiallattice}.
 
We now introduce the events corresponding to the price movements. We will see that the information embedded in the filtrations becomes increasingly detailed and specific over time. At $t=0$, all paths are possible. Thus, the event set $A=\{uuu, uud, udu, udd, ddd, ddu, dud, duu\}$ -- with the sequences describing the movement per time step -- contains all possible paths $\omega \in \Omega$, such that $A\equiv\Omega$. At $t=1$, we know that the stock price went either up or down. The corresponding events can be defined by $A_u=\{uuu,uud,udu,udd\}$ and $A_d=\{ddd,ddu,dud,duu\}$. Note that if the price went up, we know our sample path $\omega$ will be in $A_u$ and not in $A_d$. At $t=2$, we have four event sets: $A_{uu}=\{uuu,uud\}$, $A_{ud}=\{udu,udd\}$, $A_{du}=\{duu,dud\}$, and $A_{dd}=\{ddu,ddd\}$. Observe that the information is getting increasingly fine-grained; the sets to which $\omega$ might belong are becoming smaller and more numerous. At $t=3$, we obviously know the exact price path that has been followed. Having defined the events, we can define the corresponding filtrations for $t=0,1,2,3$:
 
  \begin{tabular}[H]{l l }
  	&\\
  	$\mathcal{F}_0=$& $\{\emptyset, \Omega\},$\\ 
  	&\\
  	$\mathcal{F}_1=$& $\{\emptyset, \Omega,A_u,A_d\}$\\
  	&\\
    	$\mathcal{F}_2=$&$\{\emptyset, \Omega, A_u, A_d, A_{uu}, A_{ud},A_{du}, A_{dd},$\\ &$ A_{uu}^C,A_{ud}^C, A_{du}^C ,A_{dd}^C,$ \\
    &	$A_{uu}\cup A_{du},	A_{uu}\cup A_{dd},$ \\
    &	$A_{ud}\cup A_{du},A_{ud}\cup A_{dd} \}$\\
  	$\mathcal{F}_3=$& all (256) subsets of $\Omega$
  \end{tabular}
  \vspace{0.5cm}
  
For $\mathcal{F}_0$, it is eminent that any $\omega$ must belong to $\Omega$ and not to $\emptyset$. We have not observed any information that allows for a more accurate classification. For $\mathcal{F}_1$, we can define two more sets to which $\omega$ may belong. Due to observing the first price change, we are now able to assign $\omega$ to $A_u$ or $A_d$; we may state that these sets are `resolved'. When moving to $\mathcal{F}_2$, things get slightly more involved. Whenever we have resolved a set, we have also resolved its complement. In $\mathcal{F}_1$ we had $A_u^C=A_d$ and vice versa, but for $\mathcal{F}_2$ we must explicitly define the complements (e.g., a path either is in $A_{uu}$ or in $A_{uu}^C$). Finally, whenever multiple sets are resolved, so is their union. Again, in $\mathcal{F}_1$ we had $A_u \cup A_d =\Omega$, so an explicit union definition was not necessary. For $\mathcal{F}_2$ however, we must explicitly include the unions, e.g., $A_{ud}\cup A_{dd}$. Note that the triple unions are equivalent to complements and that the quadruple union equals the outcome space.

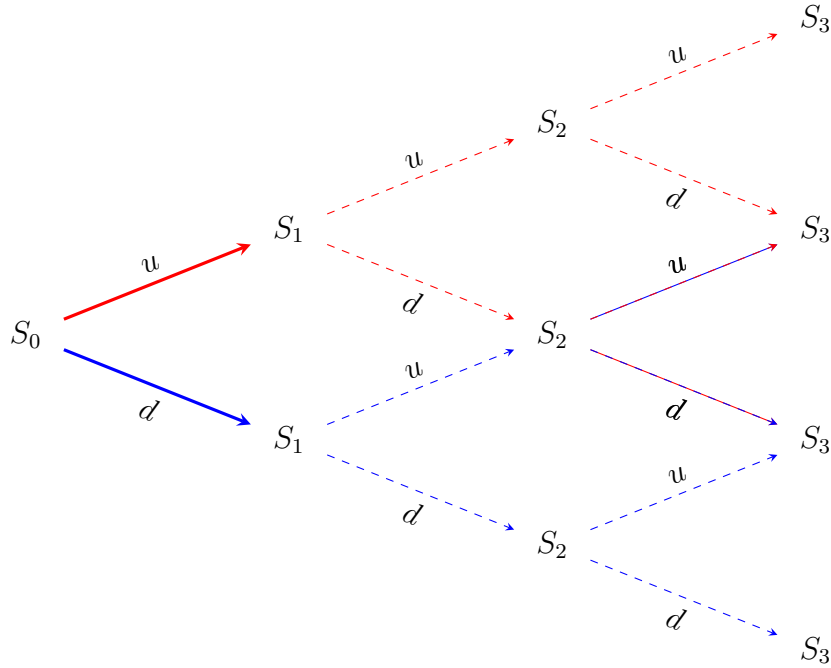
\begin{figure}[H]
  	\centering
   \begin{tikzpicture}[>=stealth,sloped]
   \matrix (tree) [%
   matrix of nodes,
   minimum size=1cm,
   column sep=2.5cm,
   row sep=0.4cm,
   ]
   {  
   	&         &         &  $S_3$             \\
   	&         &  $S_2$  &                  \\
   	& $S_1$   &         &  $S_3$             \\
   	$ S_0$ &         &  $S_2$  &                  \\
   	& $S_1$   &         &  $S_3$         \\
   	&         &  $S_2$  &                \\
   	&         &         &  $S_3$    \\
   };
   \draw[->,very thick,red] (tree-4-1) -- (tree-3-2) node [midway,above,black] {$u$};
   \draw[->,very thick,blue] (tree-4-1) -- (tree-5-2) node [midway,below,black] {$d$}; 
   \draw[->,red,dashed] (tree-3-2) -- (tree-2-3) node [midway,above,black] {$u$};
   \draw[->,red,dashed] (tree-3-2) -- (tree-4-3) node [midway,below,black] {$d$};
   \draw[->,blue,dashed] (tree-5-2) -- (tree-4-3) node [midway,above,black] {$u$};
   \draw[->,blue,dashed] (tree-5-2) -- (tree-6-3) node [midway,below,black] {$d$};
   
    \draw[->,blue] (tree-4-3) -- (tree-3-4) node [midway,above,black] {$u$};
    \draw[->,red] (tree-4-3) -- (tree-5-4) node [midway,below,black] {$d$};
   
   \draw[->,red,dashed] (tree-2-3) -- (tree-1-4) node [midway,above,black] {$u$};
   \draw[->,red,dashed] (tree-2-3) -- (tree-3-4) node [midway,below,black] {$d$};
   \draw[->,red,dashed] (tree-4-3) -- (tree-3-4) node [midway,above,black] {$u$};
   \draw[->,blue,dashed] (tree-4-3) -- (tree-5-4) node [midway,below,black] {$d$};
   \draw[->,blue,dashed] (tree-6-3) -- (tree-5-4) node [midway,above,black] {$u$};
   \draw[->,blue,dashed] (tree-6-3) -- (tree-7-4) node [midway,below,black] {$d$};

   \end{tikzpicture}
     \caption{Intuitive visualization of $\mathcal{F}_1$. The colors red and blue indicate the event sets $A_u$ and $A_d$ after observing one stock price movement.}
     \label{fig:filtration1}
   \end{figure}

From this example it can be seen that $\mathcal{F}_t\subseteq\mathcal{F}_{t+1}$ indeed holds. The filtration at time $t$ embeds all event sets that can be distinguished up until that point, based on the possible realizations of the random variables $W_1,\ldots,W_t$. We further illustrate this result with some figures. Figure~\ref{fig:filtration1} visualizes the event sets $A_u$ and $A_d$:

%The filtration $\mathcal{F}_1$ can be visualized as shown in Figure~\ref{fig:filtration1}, including the initial state $S_0$ and two price paths, up and down. Although we cannot observe the future, information from . For instance, if the first price movement is `up', we know we will never reach the lowest price at $S_3$. 
 
The filtration $\mathcal{F}_2$ encapsulates $\mathcal{F}_1$ and also takes into account the return information revealed at $t=2$. Thus, we now has event sets $A_{uu}$, $A_{ud}$, $A_{du}$ and $A_{dd}$, illustrated by the distinct colors in Figure~\ref{fig:filtration2}:
 
 \begin{figure}[H]
 	\centering
  \begin{tikzpicture}[>=stealth,sloped]
  \matrix (tree) [%
  matrix of nodes,
  minimum size=1cm,
  column sep=2.5cm,
  row sep=0.4cm,
  ]
    {  
    	&         &         &  $S_3$             \\
    	&         &  $S_2$  &                  \\
    	& $S_1$   &         &  $S_3$             \\
    	$ S_0$ &         &  $S_2$  &                  \\
    	& $S_1$   &         &  $S_3$         \\
    	&         &  $S_2$  &                \\
    	&         &         &  $S_3$    \\
    };
  
  \draw[->,very thick,red] (tree-4-1) -- (tree-3-2) node [midway,above,black] {$u$};
  \draw[->,very thick,dotted,green] (tree-4-1) -- (tree-3-2) node [midway,above,black] {$u$};
  \draw[->,very thick,blue] (tree-4-1) -- (tree-5-2) node [midway,below,black] {$d$};
  \draw[->,very thick,dotted,orange] (tree-4-1) -- (tree-5-2) node [midway,below,black] {$d$};
  \draw[->,very thick,red] (tree-3-2) -- (tree-2-3) node [midway,above,black] {$u$};
  \draw[->,very thick,green] (tree-3-2) -- (tree-4-3) node [midway,below,black] {$d$};
  \draw[->,very thick,orange] (tree-5-2) -- (tree-4-3) node [midway,above,black] {$u$};
  \draw[->,very thick,blue] (tree-5-2) -- (tree-6-3) node [midway,below,black] {$d$};
  
        \draw[->,green] (tree-4-3) -- (tree-3-4) node [midway,above,black] {$u$};
        \draw[->,orange] (tree-4-3) -- (tree-5-4) node [midway,below,black] {$d$};

   \draw[->,red,dashed] (tree-2-3) -- (tree-1-4) node [midway,above,black] {$u$};
   \draw[->,red,dashed] (tree-2-3) -- (tree-3-4) node [midway,below,black] {$d$};
   \draw[->,orange,dashed] (tree-4-3) -- (tree-3-4) node [midway,above,black] {$u$};
   \draw[->,green,dashed] (tree-4-3) -- (tree-5-4) node [midway,below,black] {$d$};
   \draw[->,blue,dashed] (tree-6-3) -- (tree-5-4) node [midway,above,black] {$u$};
   \draw[->,blue,dashed] (tree-6-3) -- (tree-7-4) node [midway,below,black] {$d$};

  \end{tikzpicture}
  \caption{Intuitive visualization of $\mathcal{F}_2$. The colors red, green, orange and blue indicate event sets $A_{uu}$, $A_{ud}$, $A_{du}$, $A_{dd}$ respectively. Note that this filtration is more fine-grained than $\mathcal{F}_1$.}
   \label{fig:filtration2}
   \end{figure}
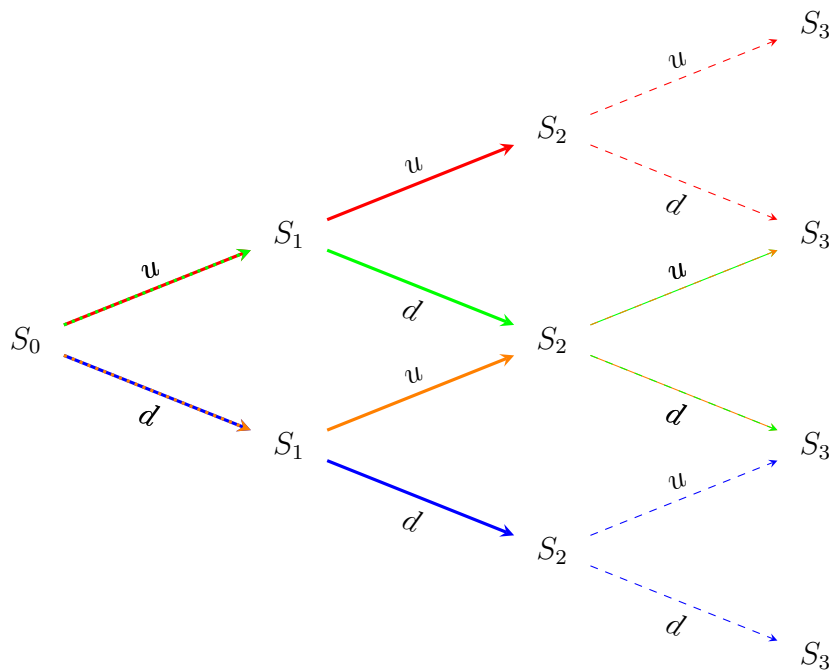
   
This tiny lattice example extends to all problems with discrete outcome spaces $\Omega$ and time horizons $T$ of any size. However, an augmentation to continuous outcome spaces is not necessarily trivial. Suppose that rather than a lattice model, we use a continuous stochastic process to  generate returns at each discrete time step \cite{vanheeswijk2012}. For the purpose of illustration, let us assume that the return may be any number in $[-d,u]\cup \mathbb{R}$, i.e., any real number between $-d$ and $u$. The outcome space $\Omega$ is now continuous. The core concept of the filtration remains unchanged for continuous outcome spaces, but requires some more attention. Again, the filtration  embeds all events based on every possible price path, the unions of these events, and the complements of these events. However, it is no longer eminent what an `event' is; individual outcomes have probability 0 in continuous space \cite{shreve2004}. Simply stated, the `event' is something we want to measure. On the real line, we often use the Borel $\sigma$-algebra, which contains all open intervals, their unions and their complements. For instance, on a domain $[S-d,S+u]\cup \mathbb{R} = [329,335]$ we could define a Borel $\sigma$-algebra $\mathcal{B}[329,335]$. Such algebras may contain intervals\footnote{As individual points have a probability of 0 occurring, open and closed sets have the same probability.} such as 

\begin{equation}
[330.3,331.9), (329.2221,332.2304), [332.50,334.64], \notag
\end{equation}

as well as all their unions and complements. The complement of $[330.3,331.9)$ would be 

\begin{equation}
[330.3,331.9)^C=[329,330.3)\cup[331.9,335]. \notag
\end{equation}

Furthermore, we can construct a plethora of unions such as 

\begin{equation}
(329.2221,332.2304)=\bigcup_{n=1}^{\infty} \left[329.2221+\frac{1}{n},332.2304-\frac{1}{n}\right]. \notag
\end{equation}

\begin{figure}[H]
 	\centering
 	\includegraphics{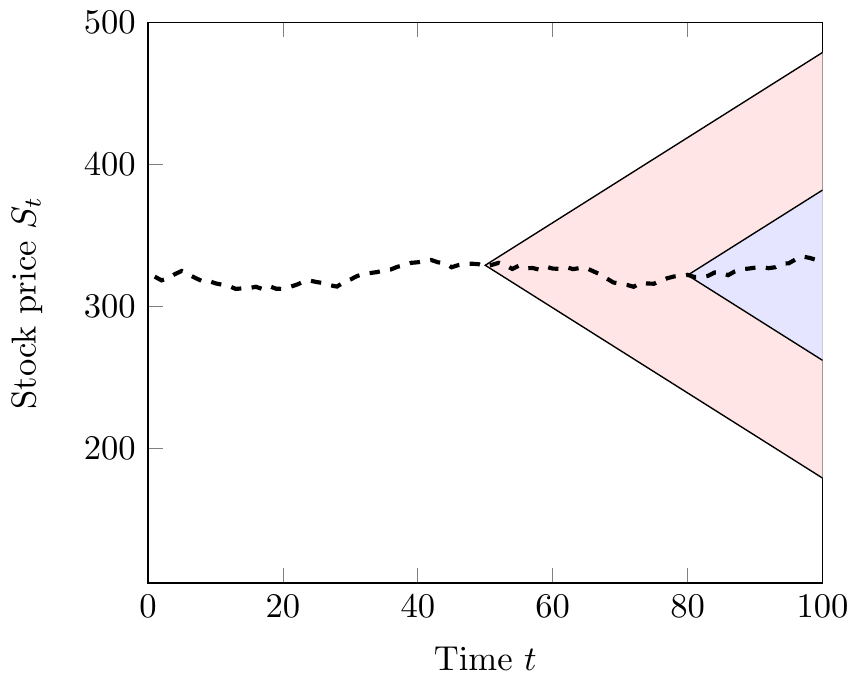}
 	\caption{Visualization of a simulated price path $\omega$ and two event sets in continuous domain. Note that the event set at $t=80$ has a finer resolution than the set at $t=50$.}
 	\label{fig:filtrationcontinuous}
 \end{figure}

Although we can think of infinitely many events, we may assign a positive probability to each of them and verify whether or not the price path is in the interval. If we start with price $S_0$, the price at $t=1$ falls within $[S_0-d,S_0+u]$, at $t=2$ it falls within $[S_0-2d,S_0+2u]$, etc. Thus, the outcome space may be visualized as a cone shape that contains all possible price paths. As time passes, we can define increasingly narrow boundaries, although within these boundaries we can define an infinite number of open intervals (and their complements and unions). Figure~\ref{fig:filtrationcontinuous} illustrates two possible event sets corresponding to a simulated price path $\omega$ in continuous space. 

%We reiterate that the filtration still only contains information up to $t$, composed of the sequence of returns realized by $W_1,\ldots,W_t$. 

%To re-iterate the key message once more, $\mathcal{F}$ is a $\sigma$-algebra on the outcome space $\Omega$ that captures all events that may occur. The sequence $\mathcal{F}_1,\ldots,\mathcal{F}_T$ is what we call a filtration. Each $\mathcal{F}_t$ describes all possible event paths up till $t$.

To wrap up, we revisit the use of filtrations in a reinforcement learning context. In each learning episode, we construct a sample path $\omega$ that is typically randomly drawn from the outcome space. If we define a decision $x_t(\omega)$ based on the outcome space, we would already know all information, including events revealed at $t+1,\ldots,T$. In our stock price example, we would know exactly when to buy or sell, having perfect insight into the price movements up to $T$. However, if we impose that $x_t(\omega)$ is $\mathcal{F}_t$-measurable, decisions can only be made based on the information up till time $t$, such that realizations of $W_{t+1},\ldots,W_T$ are not taken into account when making a decision at $t$. This way, the notion of filtrations elegantly resolves the issue of prematurely revealing future information to the decision maker.

Recall that in RL, we aim to find a decision-making policy $\pi: S_t \mapsto x_t$. The state $S_t$ can be computed based on the initial state $S_0$, the decisions made, and the information sequence $W_1=\omega_1,\ldots,W_t=\omega_t$. However, as the Markovian property holds (remind that decisions only depend on the current state  of the system, not on information from the past), we need solely our current state $S_t$ to make a decision, not the entire information sequence leading to that state. In case of our stock price example, decisions whether to sell or buy only depend on the current stock price, which implicitly embeds all price fluctuations of the past. Hence, when stripping our MDP model to the minimum information that is strictly necessary to make a decision, the notion of filtrations is redundant. Nevertheless, filtrations are generic and broadly applicable, which is why many authors opt to use filtration concept in the formal definition of their MDPs. Ultimately, it boils down to convention and background. Whether utilizing the concept or not, for anyone active in the RL domain it is useful to have at least an intuitive understanding of the concept of filtrations.

 \bibliographystyle{apalike}
 \bibliography{bibliographyfiltration}

\end{document}